\documentclass[letterpaper, 10 pt, conference]{ieeeconf}  

\IEEEoverridecommandlockouts
\usepackage[table]{xcolor}
\usepackage{cite}

\usepackage{amsmath,amssymb,amsfonts}
\usepackage{graphicx}
\usepackage{textcomp}
\usepackage{authblk}
\let\labelindent\relax
\usepackage{enumitem}
\usepackage{breqn}
\usepackage{algpseudocode}
\usepackage[ruled,lined]{algorithm2e}

\usepackage{subfigure}

\usepackage{array}
\usepackage{makecell}
\usepackage{tabularray}
\usepackage{bm}

\usepackage{svg}
\usepackage{booktabs}
\usepackage{multirow}
\usepackage[colorlinks,linkcolor=red,anchorcolor=blue,citecolor=green]{hyperref}
\usepackage{threeparttable}
\usepackage{algorithm}

\title{\LARGE \bf
JAM: Keypoint-Guided Joint Prediction after Classification-Aware Marginal Proposal for Multi-Agent Interaction}
\author{
Fangze Lin$^{1,\alpha}$, Ying He$^{1,\star}$, Fei Yu$^{1,2}$ and Hong Zhang$^{3}$
\thanks{This work is supported in part by Shenzhen Science and Technology Program under Grant ZDSYS20220527171400002, the National Natural Science Foundation of China (NSFC) under Grants 62271324, 62231020 and 62371309, the Guangdong Basic and Applied Basic Research Foundation under Grant 2023A1515011979, the Hetao Shenzhen-HongKong Science and Technology Innovation Cooperation Zone (HZQB-KCZYZ-2021055), and Shenzhen Deeproute.ai Co., Ltd (HZQB-KCZYZ-2021055).}
\thanks{$^1$College of Computer Science and Software Engineering, Shenzhen University, P.R. China. (\texttt{linfangze2023@email.szu.edu.cn}, \texttt{heying@szu.edu.cn})} 
\thanks{$^2$Carleton University, Canada. (\texttt{yufei@szu.edu.cn})}
\thanks{$^3$Shenzhen Key Laboratory of Robotics and Computer Vision, SUSTech, Shenzhen, China. (\texttt{hzhang@sustech.edu.cn})}
\thanks{$^\star$Corresponding Author: Ying He.}
\thanks{$^\alpha$Work done during an internship at Carizon.}
}

\begin{document}
\maketitle
\thispagestyle{empty}
\pagestyle{empty}
\begin{abstract}
Predicting the future motion of road participants is a critical task in autonomous driving. In this work, we address the challenge of low-quality generation of low-probability modes in multi-agent joint prediction. To tackle this issue, we propose a two-stage multi-agent interactive prediction framework named \textit{keypoint-guided joint prediction after classification-aware marginal proposal} (JAM). The first stage is modeled as a marginal prediction process, which classifies queries by trajectory type to encourage the model to learn all categories of trajectories, providing comprehensive mode information for the joint prediction module. The second stage is modeled as a joint prediction process, which takes the scene context and the marginal proposals from the first stage as inputs to learn the final joint distribution. We explicitly introduce key waypoints to guide the joint prediction module in better capturing and leveraging the critical information from the initial predicted trajectories. We conduct extensive experiments on the real-world Waymo Open Motion Dataset interactive prediction benchmark. The results show that our approach achieves competitive performance. In particular, in the framework comparison experiments, the proposed JAM outperforms other prediction frameworks and achieves state-of-the-art performance in interactive trajectory prediction. The code is available at https://github.com/LinFunster/JAM to facilitate future research.

\end{abstract}
\section{Introduction}
\begin{figure}[t]
    \setlength{\abovecaptionskip}{0cm}

    \centering
    \includegraphics[width=1.0\columnwidth]{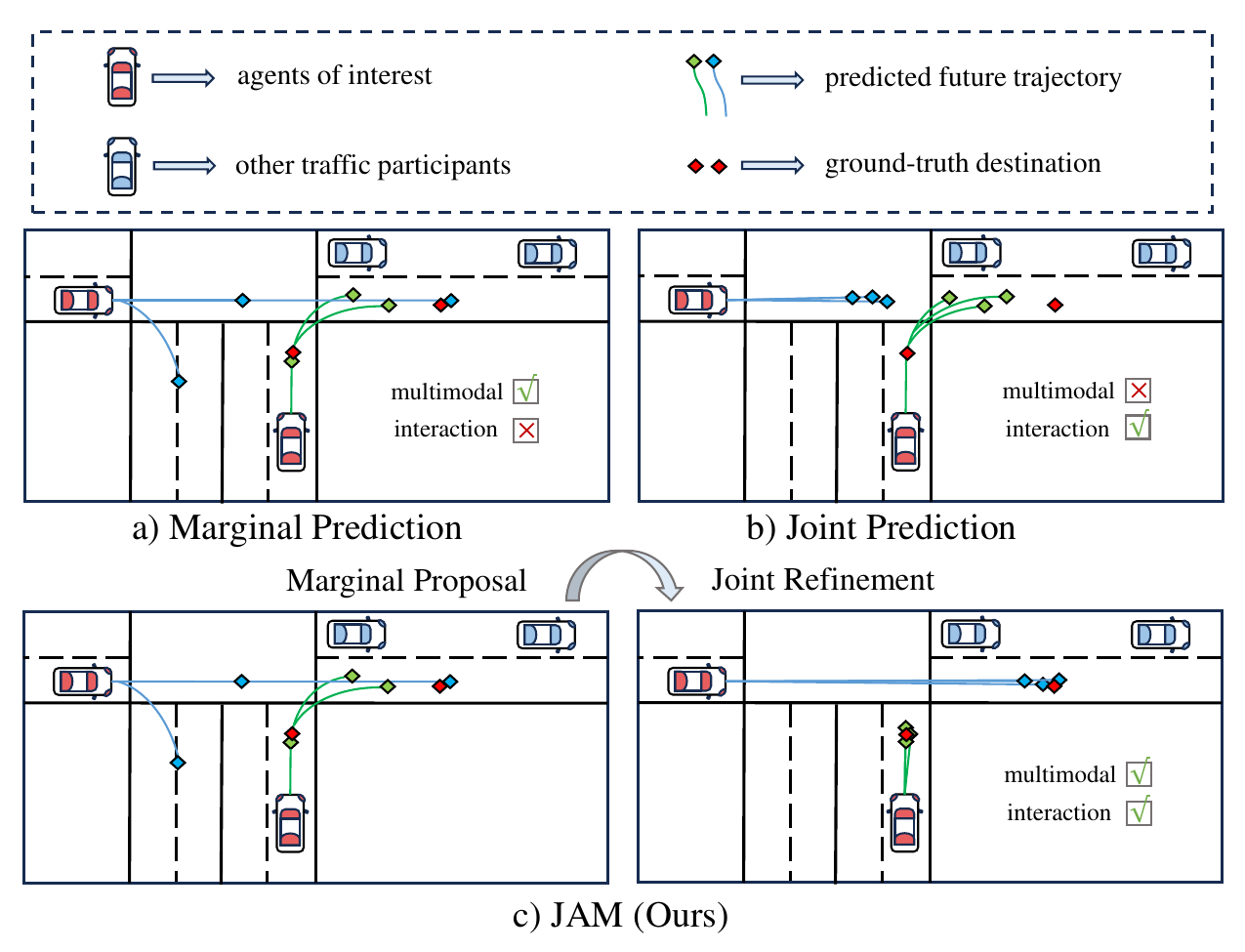}
    \vspace{-0.6cm}
    \caption{Three different motion prediction paradigms. 
    \textbf{(a)} Marginal prediction. 
    \textbf{(b)} Joint prediction. 
    \textbf{(c)} Our proposed method.
    }
    \label{intro_figure}
    \vspace{-0.3cm}
\end{figure}

Multi-agent motion prediction is a crucial task in the autonomous driving pipeline, widely used to infer the future movements of multiple road participants in complex driving environments, enabling proactive avoidance of hazardous driving actions. Most existing multi-agent motion prediction approaches predict a set of marginal trajectories for each agent \cite{jia2023hdgt,liang2020learning,liu2021multimodal,varadarajan2022multipath++,ye2021tpcn,gao2020vectornet, gu2021densetnt}. While marginal prediction methods have achieved significant success in modeling the motion of individual agents, they fail to accurately predict the interactions between multiple agents in the future. As illustrated in Figure \ref{intro_figure}, the most likely marginal predictions for two interacting agents may lead to a collision. Given that consistent future predictions are critical for downstream planning, recent works have shifted towards generating a set of scene-level or joint future trajectory predictions \cite{casas2020implicit,chen2022scept,cui2021lookout,ngiam2021scene,sun2022m2i}, where each mode consists of the joint future trajectories of multiple agents.

We focus on the problem of generating a set of joint future trajectory predictions in multi-agent driving scenarios. Unlike marginal prediction, the joint trajectory prediction space grows exponentially with the number of agents in the scene, making this prediction setup particularly challenging. A common approach in this setting is to simultaneously predict the joint future trajectories of all agents in the scene \cite{casas2020implicit,cui2021lookout,girgis2021latent,huang2023gameformer, zhou2023qcnext,ngiam2021scene}. However, directly learning the joint future of k agents often fails to effectively capture the diverse modalities of each agent's future trajectories. \cite{cui2019multimodal,liang2020learning,ngiam2021scene,varadarajan2022multipath++,huang2023gameformer, zhou2023qcnext}.

To address this, we propose the Keypoint-Guided \textbf{J}oint Prediction \textbf{a}fter Classification-Aware \textbf{M}arginal Proposal (JAM) framework. In the first stage, we introduce classification-aware mode queries that enforce the model to predict marginal trajectories covering all categories, where each marginal trajectory and its corresponding keypoints together form a proposal. This enables the prediction model to explicitly learn and consider the diverse future modalities of each agent’s trajectory. Additionally, since each proposal is learnable, this avoids the reliance on handcrafted proposal quality, which could otherwise negatively impact model performance \cite{deo2018convolutional,chai2019multipath,phan2020covernet,zeng2021lanercnn,zhao2021tnt}. The second stage is modeled as a joint prediction process that takes scene context and marginal proposals information as input to generate joint trajectory predictions.

We conduct extensive experiments on the real-world Waymo Open Motion Dataset interactive prediction benchmark, demonstrating competitive performance. In the framework comparison experiments, the proposed JAM outperforms other prediction frameworks, achieving state-of-the-art interactive prediction results. Additionally, we perform ablation studies to validate the effectiveness of each component and the impact of various parameter settings.

Our main contributions can be summarized as follows:
\begin{itemize}
    \item We propose the JAM framework, which simultaneously benefits from marginal prediction for multi-modal modeling and joint prediction for interaction modeling.
    \item We propose the classification-aware marginal proposal method, which classifies mode queries to enable the joint prediction model to explicitly learn and consider the diverse modalities of each agent’s future trajectories.
    \item We propose keypoint-guided encoding, which explicitly incorporates keypoints from marginal trajectories to help the model better capture critical information in future trajectories.
\end{itemize}

\section{Related Work}
\subsection{Joint Motion Prediction}
Most existing motion prediction systems generate marginal predictions for each agent \cite{jia2023hdgt,liang2020learning,liu2021multimodal,varadarajan2022multipath++,ye2021tpcn,gao2020vectornet,gu2021densetnt,shi2022motion,zeng2021lanercnn,zhou2022hivt}. However, marginal predictions lack future associations across agents. Recent research has explored generating joint predictions simultaneously \cite{cui2021lookout,casas2020implicit,girgis2021latent,huang2023gameformer, zhou2023qcnext,ngiam2021scene}. However, freely generating multiple most-probable hypotheses often leads to modal omissions and training instability \cite{cui2019multimodal,liang2020learning,ngiam2021scene,varadarajan2022multipath++,huang2023gameformer, zhou2023qcnext}. Additionally, some works have started considering generating joint predictions for two-agent interaction scenarios by selecting $K$ joint futures from all possible $K^2$ combinations of marginal predictions \cite{shi2022motion,wu2021air}. Other works adopt conditional methods, predicting the motion of other agents based on the movement of a controlled agent \cite{rowe2023fjmp, sun2022m2i}. However, effectively combining marginal and joint predictions remains a challenging problem. To address this, the proposed JAM framework first performs marginal prediction to generate classification-aware proposals, followed by a keypoint-guided joint prediction process. JAM effectively combines the strengths of both marginal and joint predictions, achieving better performance than any single framework.

\subsection{Interaction Modeling for Motion Prediction}
Data-driven methods typically use attention-based mechanisms \cite{liang2020learning,liu2021multimodal,girgis2021latent,ngiam2021scene,zhou2022hivt} or graph neural networks (GNNs) \cite{jia2023hdgt,casas2020implicit,gao2020vectornet,kumar2021interaction,zeng2021lanercnn} to model interactions between agents for motion prediction. Recent research has started to focus on predicting potential agent interactions in the future \cite{kuo2022trajectory,huang2023gameformer,sun2022m2i, shi2024mtr++, zhou2023qcnext}. \cite{shi2022motion,shi2024mtr++} propose generating future trajectory intentions as an auxiliary task, where the future intentions are fed into an interaction module for better reasoning about future interactions. Some works infer future interactions through predicted influencer-reactor relationships \cite{rowe2023fjmp,sun2022m2i}, where the agent arriving at the conflict point first is defined as the influencer, and the other as the reactor. Gameformer \cite{huang2023gameformer}, on the other hand, employs hierarchical game theory to frame the interaction prediction problem, modeling complex game-theoretic interactions. In contrast, JAM leverages the attention mechanism to model the agent's historical interactions and employs a two-stage interactive prediction framework to capture future trajectory interactions.

\subsection{Two-Stage Trajectory Prediction}
Two-stage decoding frameworks have become increasingly common in recent trajectory prediction methods \cite{zhou2023query, zhou2023qcnext, shi2022motion, shi2024mtr++, choi2023r, huang2023gameformer, zhou2024smartrefine}. In the first stage, trajectory proposals are generated and used as inputs for the second stage. The second stage then outputs either the offsets for each proposal trajectory or predicts the final trajectories. There are two main designs for the second stage: one-step re-prediction and multi-step re-prediction. One-step re-prediction typically employs larger and more complex decoding modules \cite{zhou2023query, zhou2023qcnext, choi2023r}. In contrast, multi-step re-prediction usually adopts lightweight and simpler decoding structures \cite{shi2022motion, shi2024mtr++, huang2023gameformer, zhou2024smartrefine}. In this work, we propose JAM, a two-stage decoding framework with one-step re-prediction. JAM combines the strengths of marginal prediction and joint prediction to further unlock the potential of two-stage decoding frameworks.
\begin{figure*}[t]
  \vspace{0.2cm}
  \centering
  \includegraphics[width=2.0\columnwidth]{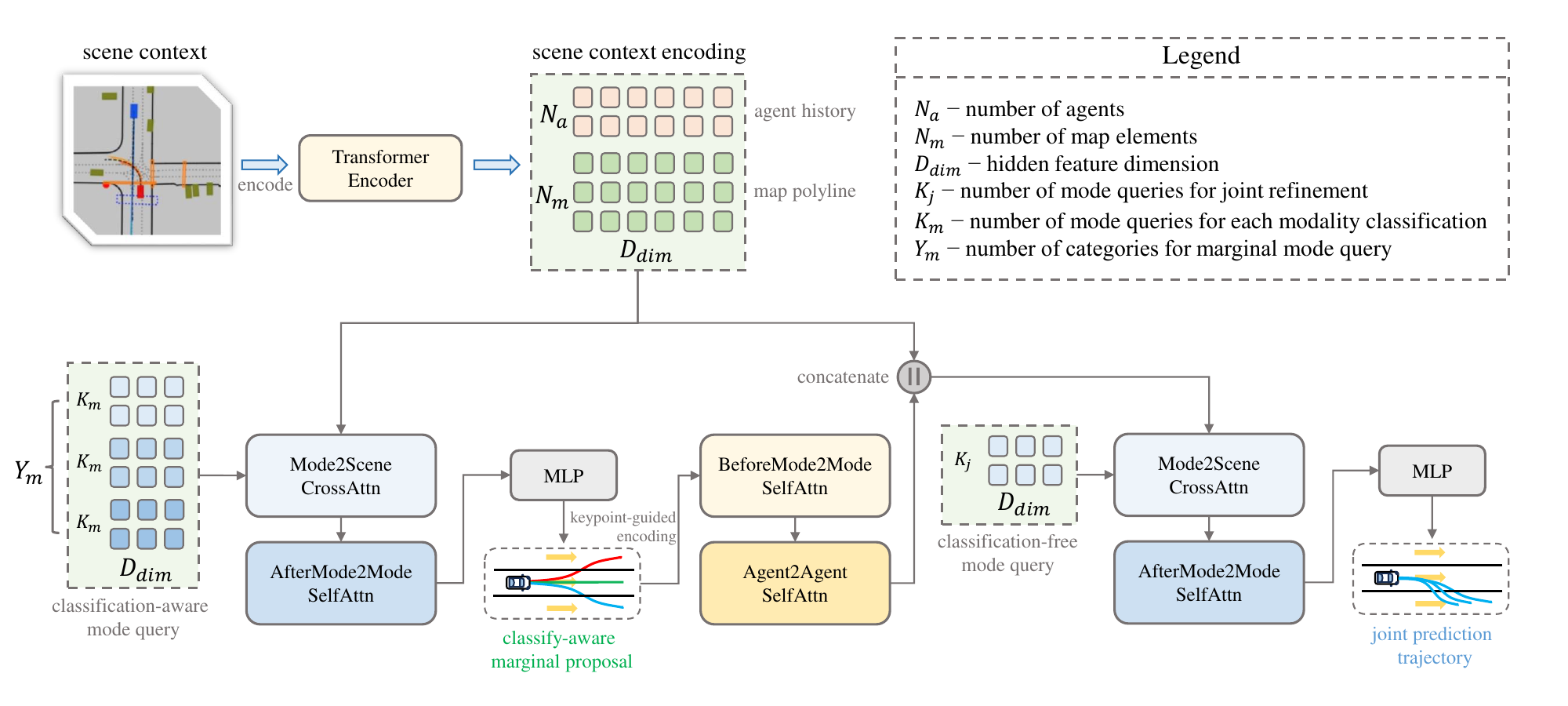}
  \vspace{-0.3cm}
  \caption{Overview of the two-stage decoding pipeline. The figure above illustrates the decoding framework of Keypoint-Guided Joint Prediction after Classification-Aware Marginal Proposal (JAM). For the encoding part, we employ a query-centric scene context encoding, where rotation and centering are applied to both input and output to facilitate convergence. The decoding part is divided into two stages. In the first stage, we utilize classification-aware marginal prediction to obtain the initial trajectories. By classifying the queries, we enforce the model to predict all category marginal trajectories, ensuring more comprehensive modality consideration. The second stage is modeled as a joint prediction process. We consider the scene context and the marginal prediction trajectories. In addition, by explicitly incorporating keypoints from the previous stage’s output trajectories, the model more easily learns the interactions among multiple agents.}
  \vspace{-0.1cm}
  \label{fig:system_overview_figure}
\end{figure*}

\section{Approach}
\subsection{Symbol Definition}
We consider a scene with $N_a$ agents, where the agent types belong to one of the categories: vehicle, bicycle, or pedestrian. For the input data, we include the historical state information of the agents, denoted as $F_p \in \mathbb{R}^{N \times T_h \times d_s}$, where $d_s$ represents the number of state attributes and $T_h$ denotes the number of historical time steps considered. The local vectorized map polyline is represented as $M \in \mathbb{R}^{N \times N_m \times N_p \times d_p}$. For each agent, we identify $N_m$ nearby map elements, including roads and crosswalks. Each map element consists of $N_p$ road points, each with $d_p$ attributes, with missing values padded with zeros in the tensor. For the output data, given the map information and agent states within an observation window of $T$ time steps, the multimodal trajectories are modeled as a Gaussian Mixture Model (GMM), where the state of the $i$-th agent at time step $t$ is represented as $F^t_i = (\mu_x, \mu_y, \sigma_x, \sigma_y)$. The task of the prediction module is to predict $K$ future trajectories for each target agent over the range of $T$ time steps and assign a probability score to each prediction. The marginal predicted trajectories in the first stage are divided into $Y_m$ categories, with $K_m$ mode queries for each marginal prediction category. Additionally, $K_j$ denotes the number of mode queries for joint prediction. The hidden layer dimension of the model's attention mechanism is $D_{dim}$.

\subsection{Query-Centric Scene Context and Future Trajectory Encoding}
Similar to \cite{shi2024mtr++, zhou2023query}, we adopt a query-centric approach for encoding the input. Both the scene context (including map elements and agent historical information) and predicted future trajectories establish a local coordinate system centered at a specific point of each element. These are then encoded via a neural network. The local coordinate origin retains information from the ego vehicle coordinate system and undergoes positional encoding. Subsequently, the element encoding and the origin encoding are added together to integrate global information. This approach maintains rotational consistency, which facilitates neural network training and convergence.

We use an LSTM network to encode the historical trajectories of each agent and employ an Multilayer Perceptron (MLP) to encode the vectorized map information. Subsequently, we utilize a Transformer encoder with $E$ layers to capture the relationships between all scene elements within the context tensor of each agent. Additionally, we use an MLP network combined with a maxpool operation to encode the marginally predicted future trajectories, and an MLP network combined with an average operation to encode keypoint information. The encoded future trajectories and keypoint features are then summed with the agent‘s type embeddings before being output.

\subsection{Classification-Aware Marginal Proposal}
In the first stage of the decoder, we employ classification-aware marginal prediction to output comprehensive modal trajectories. We conduct experiments using two classification methods. The first method divides the trajectories into eight categories: stationary, straight, straight left, straight right, left turn, right turn, left U-turn, and right U-turn \cite{ettinger2021large}. Each category has $K_m$ adaptive mode queries. The second method uses $64$ anchors obtained via k-means clustering to perform region segmentation \cite{shi2022motion}. For the learnable mode query embeddings, we aggregate the following information: the modality index embedding, the agent index embedding, and the fused agent historical information encoded feature. We input the mode query embeddings of dimension $[K_mY_m, D_{dim}]$ into the Mode2Scene cross-attention layer to update the mode queries with multiple contexts, including the historical encodings of the multi-agent agents and map encodings. Then, the mode queries of dimension $[K_j, D_{dim}]$ are passed through the AfterMode2Mode self-attention layer to enhance the diversity of the multi-modal predictions. Finally, the mode queries are fed into the GMM Predictor to predict the multi-modal trajectories.

\subsection{Keypoint-Guided Joint Prediction}
For joint prediction, the key to the joint decoder lies in learning the final joint distribution based on the marginal proposals from the previous stage. A marginal proposal includes a predicted trajectory generated without considering interactions with other agents, along with its corresponding keypoints. We re-encode the proposals by inputting the agent embedding with dimensions $[N_aK_mY_m, D_{dim}]$ into the BeforeMode2Mode self-attention layer to account for the interactions between multi-agent modes. Then, we take the average along the $K_mY_m$ dimensions, transforming the encoding dimensions to $[N_a, D_{dim}]$, which is subsequently fed into the Agent2Agent self-attention layer to handle interactions between agents. For the learnable initial mode query embeddings, we aggregate the following information: the modality index embedding, the agent index embedding, the fused agent historical information encoded feature, the content feature of the marginal prediction, and the encoded feature of the marginal proposal from the previous stage. We input the mode query embeddings of dimension $[K_j, D_{dim}]$ as $Q$ and the scene context as $KV$ into the Mode2Scene cross-attention layer to update the mode queries with multiple contexts, including the historical encodings of multi-agent agents, map encodings, and future interaction encodings of neighboring agents. The mode queries of dimension $[K_j, D_{dim}]$ are then passed through the AfterMode2Mode self-attention layer to improve the diversity of the multi-modal predictions. Finally, the mode queries are input into the GMM Predictor to predict the multimodal trajectories. We also propose explicitly modeling key points as inputs to the model. 

Additionally, we explicitly model keypoints to better capture critical information within the marginal predictions. We define keypoints as important waypoints along the initial prediction, where each point contains position and velocity information. In this work, to enable the model to focus on both short-term and long-term destinations, we select the predicted waypoints at 3s, 5s, and 8s as keypoints. It is important to note that the choice of keypoints is not unique. Other options include collision points, interaction points, and others. However, a detailed discussion of additional keypoints is beyond the scope of this paper and is left for future research.

\subsection{Training Objectives}
Similar to \cite{huang2023gameformer}, we use Gaussian regression loss to maximize the likelihood of the ground-truth trajectories and employ cross-entropy loss to learn the optimal modality selection. Specifically, the future behavior of an agent is modeled as a Gaussian Mixture Model (GMM), where each mode \( k \) at time step \( t \) is described by a Gaussian distribution over the (x, y) coordinates, characterized by the mean \( \mu_t \) and covariance \( \sigma_t \). The Gaussian regression loss is computed using the negative log-likelihood loss of the best-predicted component \( m^\star \) (the one closest to the ground truth), as follows:
\begin{equation}
\begin{aligned}
\mathcal{L}_{NLL} = \sum_{y=1}^{Y}\mathbb{I}(y=y_{gt})[\sum_{k_y=1}^{K_y}\mathbb{I}(k_y=k_y^{gt})[\log \sigma_x + \log \sigma_y + \\
\frac{1}{2}\left((\frac{d_x}{\sigma_x})^2+(\frac{d_y}{\sigma_y})^2\right) - \log(p_{m^*})
]]
\end{aligned}\label{eq:gmm_loss}
\end{equation}
where $d_x = s_x - \mu_x$ and $d_y = s_y - \mu_y$, $(s_x, s_y)$ is ground truth position;  $p_m$ is the probability of the selected component, and we use the cross-entropy loss in practice. 
Let \( \mathbb{I} \) be the indicator function. \( Y \) denotes the number of modality categories for the output trajectories, and \( K_y \) represents the number of output trajectories for a particular modality category. Therefore, the total number of output trajectories is \( \sum_{y=1}^{Y}{K_y} \). Additionally, \( y_{gt} \) is the index of the ground-truth future trajectory modality category of the agent, and \( k_y^{gt} \) is the index of the best prediction of the multi-agent future trajectory under a specific modality (the one that minimizes the total future displacement error across all agents). Notably, in this work, when implementing joint prediction, no modality categories are explicitly defined, so \( Y \) is set to 1.

\section{Experiments}
\subsection{Experiment Settings}
\subsubsection{Dataset and Metrics}
The model is trained and evaluated using the large-scale real-world driving dataset, the Waymo Open Motion Dataset (WOMD) \cite{ettinger2021large}, specifically designed to address the task of predicting the joint trajectories of two interacting agents. We adopt the WOMD interactive prediction setup, where the model predicts the joint future positions of two interacting agents over a time horizon of 8 seconds. The model considers adjacent agents within the scene as background information but predicts the joint future trajectory of only the two labeled interacting agents. The model is trained on the entire WOMD training dataset and evaluated on the validation set. The metrics for this task include minimum average displacement error (minADE), minimum final displacement error (minFDE), miss rate (Miss Rate), mean average precision (mAP), and soft mean average precision (Soft mAP).

\subsubsection{Implementation Details}
We perform training and evaluation on 4 RTX 4090 GPUs. All models are trained for 30 epochs on the training dataset. The total batch size is set to 256, and the learning rate for the neural network is initialized to 1e-4, with the learning rate halved every two epochs starting from the 20th epoch. In the proposal stage, the number of mode query categories is set to $Y_m=64$, with each category having $K_m=1$ mode query. In the joint prediction stage, the number of joint prediction mode query categories is set to $Y_j=1$, with each category having $K_j=6$ mode queries. The model considers 32 neighboring agents around the two interacting agents but generates trajectories only for the labeled interacting agents. We utilize a stack of $E=6$ Transformer encoder layers and set the hidden feature dimension to $D_{dim}=256$.

\begin{table*}[htp!]
\centering
\vspace{0.2cm}
\renewcommand{\arraystretch}{1.2}
\caption{comparison with baseline models on the WOMD interaction prediction test benchmark}
\scalebox{1.0}{
\setlength{\tabcolsep}{5.0pt}{
\begin{threeparttable}
\begin{tabular}{l|cccc|cccc}
\toprule
\multirow{2}{*}{\makecell[c]{Method}}&
\multicolumn{4}{c|}{Vehicle(Avg)}&\multicolumn{4}{c}{All(Avg)}\\
&
minADE $(\downarrow)$ &
minFDE $(\downarrow)$ &
Miss Rate $(\downarrow)$ &
mAP $(\uparrow)$ &
minADE $(\downarrow)$ &
minFDE $(\downarrow)$ &
Miss Rate $(\downarrow)$ &
mAP $(\uparrow)$ \\
\hline
HeatIRm4 \cite{mo2022multi}& 1.6348& 3.8579& 0.6788& 0.1405& 1.4197& 3.2595& 0.7224& 0.0844\\
SceneTransformer \cite{ngiam2021scene}& 1.0060& 2.2466& 0.3953& 0.1705& 0.9774& 2.1892& 0.4942& 0.1192\\
DenseTNT \cite{gu2021densetnt}& 1.1906& 2.6368& 0.4396& 0.2590& 1.1417& 2.4904& 0.5350& 0.1647\\
M2i \cite{sun2022m2i}& 1.4442& 3.0613& 0.4858& 0.2125& 1.3506& 2.8325& 0.5538& 0.1239\\
GameFormer \cite{huang2023gameformer} & 0.9822& 2.0745& 0.3785& 0.1856& 0.9161& 1.9373& 0.4531& 0.1376\\
MTR++ \cite{shi2024mtr++}& 0.9309& 2.0524& \textbf{0.3462}& 0.3303& 0.8795& 1.9509& \textbf{0.4143}& 0.2326\\
BeTop \cite{liu2024betop}& 1.0216& 2.3970& 0.3738& \textbf{0.3308}& 0.9744& 2.2744& 0.4355& \textbf{0.2412}\\
\hline
JAM ($Y_m$=64,$K_m$=1) & \textbf{0.9108}& \textbf{1.9795}& 0.3563& 0.2019& \textbf{0.8673}& \textbf{1.9073}& 0.4456& 0.1476\\  
\bottomrule
\end{tabular}
\begin{tablenotes}
\footnotesize
\item The All(Avg) metrics are averaged over different object types (vehicle, pedestrian, and cyclist) and evaluation times (3, 5, and 8 seconds). Additionally, Vehicle(Avg) only takes the results from the vehicle object types.
\end{tablenotes}
\end{threeparttable}
    }
}
\label{table:main_results}
\end{table*}

\begin{figure*}
\centering
\subfigure[Marginal Prediction (classification-aware)]
{\includegraphics[width=2.4in]{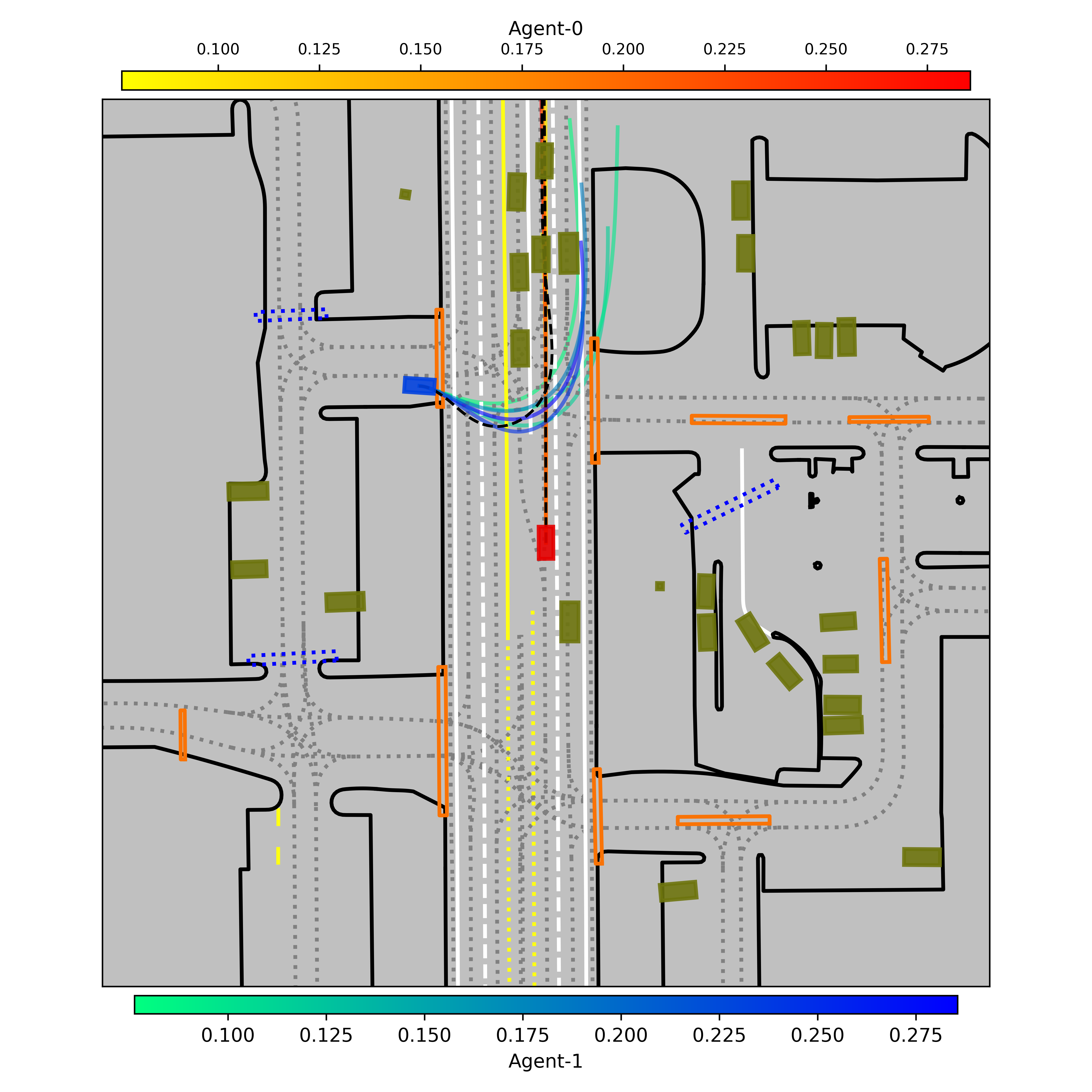}}
\hspace{-2.0em}
\subfigure[Joint Prediction (GameFormer \cite{huang2023gameformer})]{\includegraphics[width=2.4in]{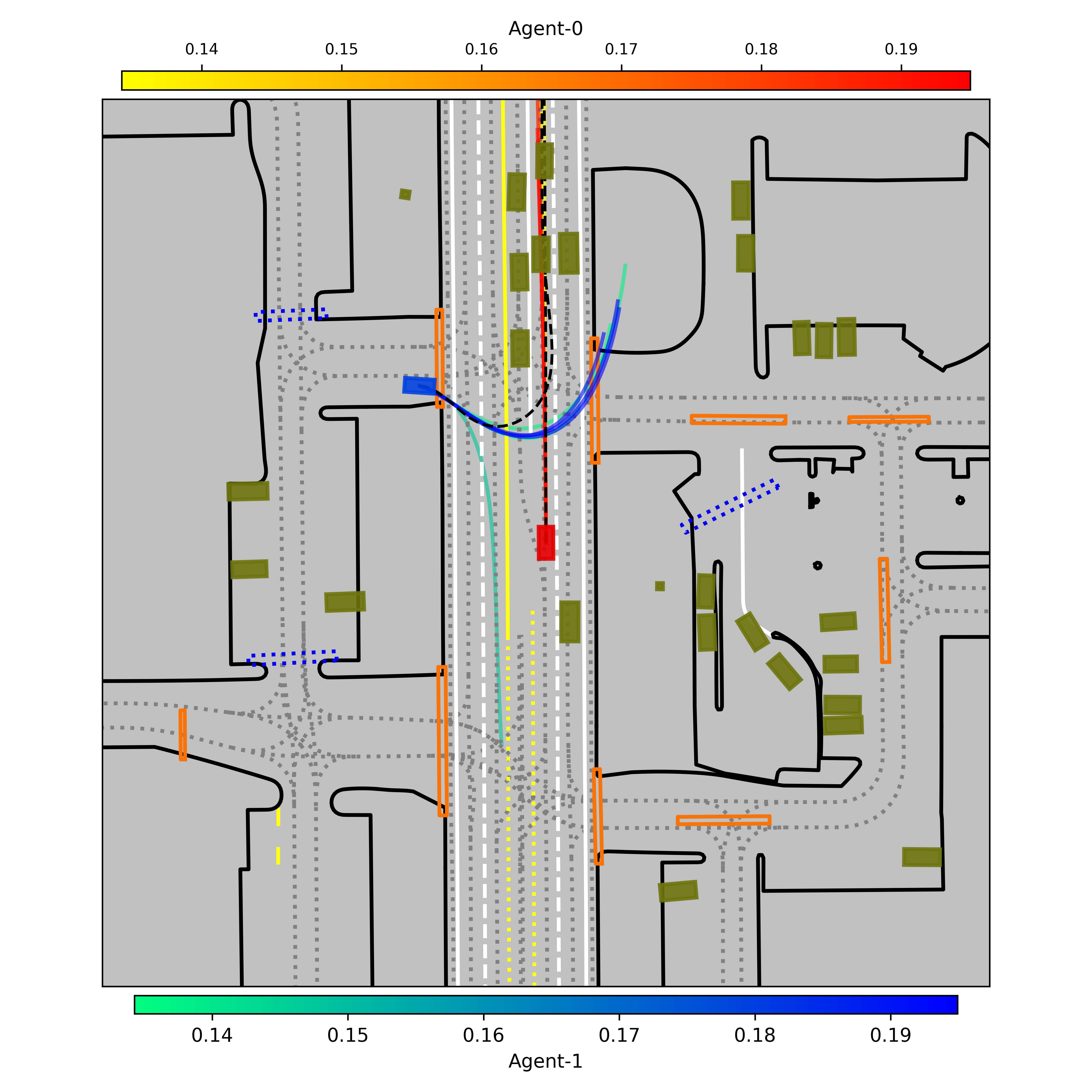}}
\hspace{-2.0em}
\subfigure[JAM ($Y_m=64,K_m=1$)]
{\includegraphics[width=2.4in]{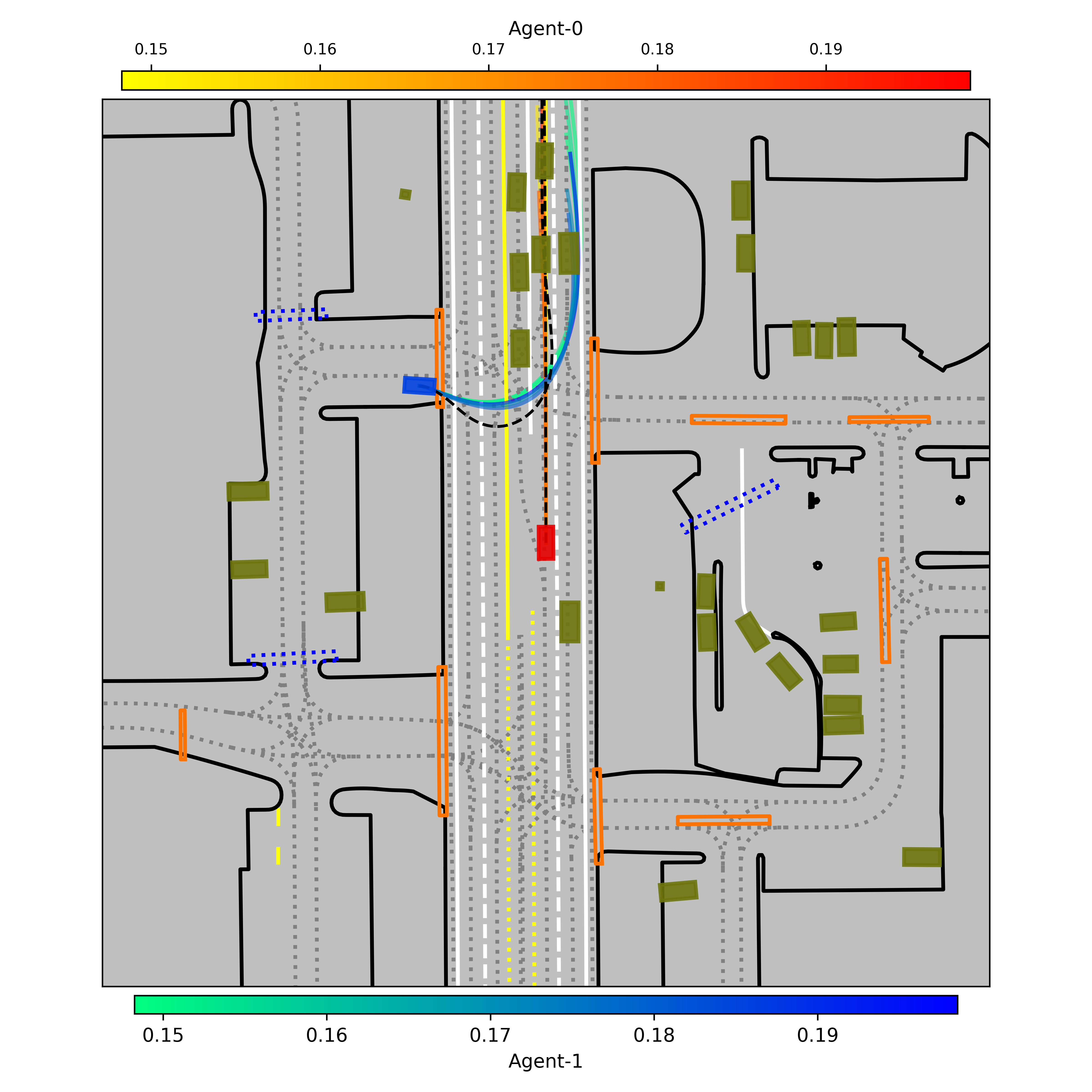}} \\
\caption{Qualitative results on the WOMD interactive validation set. The figure shows the top six highest-probability predicted trajectories, with darker colors indicating higher probabilities. Different colored filled rectangles represent different entities: \textcolor[RGB]{255,0,0}{Predicted Agent-0}, \textcolor[RGB]{0,0,255}{Predicted Agent-1}, and \textcolor[RGB]{128,128,0}{Other Vehicles}. The blue dashed rectangle represents \textcolor[RGB]{0,0,255}{Pedestrian Crosswalks}. The black dashed line indicates the \textbf{Ground-Truth Trajectories} of the predicted vehicles. Hollow circles represent \textcolor[RGB]{255,0,0}{Traffic} \textcolor[RGB]{0,128,0}{Lights}, while solid red circles indicate \textcolor[RGB]{255,0,0}{Stop Signs}.
} 
\label{fig:visual_framework_cmp}
\end{figure*}

\subsection{Baseline Comparisons for Interaction Prediction}
As shown in Table \ref{table:main_results}, we compare various baseline methods on the WOMD interaction prediction benchmark (joint prediction of two interacting agents). Our two-stage prediction model, JAM ($Y_m=64, K_m=1$), outperforms all baseline methods on minADE and minFDE metrics, while also demonstrating strong performance in terms of Miss Rate and mAP. It should be noted that our results are obtained using a single prediction model without employing any ensemble techniques. The marginal proposals used in JAM allow the model to consider more diverse modal information, which helps it handle interactions more effectively. Additionally, we present the results for Vehicle (Avg) to evaluate the model's performance in predicting the planning trajectories of other vehicles and the ego vehicle. The results indicate that JAM achieves competitive results through a simple framework, particularly excelling in position error metrics.

\begin{table}[!htp] 
\centering
\caption{comparison of result for different framework}
\vspace{-0.2cm}
\scalebox{0.58}{
\renewcommand{\arraystretch}{1.20}
\setlength{\tabcolsep}{3.0pt}{
\begin{threeparttable}
\begin{tabular}{l|cc|ccccc}
\toprule
\makecell[c]{Method} &
\makecell[c]{\#Param.} &
\makecell[c]{Inference\\Latency} &
minADE $(\downarrow)$ &
minFDE $(\downarrow)$ &
Miss Rate $(\downarrow)$ &
mAP $(\uparrow)$ &
Soft mAP $(\uparrow)$ 
\\
\hline
Marginal (classification-free)& 8.6M&11.0ms& 1.4439& 3.4907& 0.6169& 0.0904& 0.1035\\
Marginal (classfication-aware)& 8.6M&11.0ms& 1.0382& 2.3750& 0.5377& \textbf{0.1644}& 0.1683\\
Joint (one-step re-prediction)& 10.5M&15.3ms& 0.9192& 2.0273& 0.4764&0.1214& 0.1423\\
Joint (GameFormer \cite{huang2023gameformer})& 13.5M&13.7ms& 0.9012& 1.9493& 0.4576&0.1306& 0.1538\\
JAM ($Y_m$=64,$K_m$=1)& 12.2M&17.8ms& \textbf{0.8695}& \textbf{1.9028}& \textbf{0.4431}& 0.1534& \textbf{0.1742}\\
\bottomrule
\end{tabular}
\begin{tablenotes}
\footnotesize
\item \#Param.(M): number of model parameters. The inference latency is tested on an NVIDIA RTX 4090 GPU.
\end{tablenotes}
\end{threeparttable}
\label{table:framework_cmp}
    }
}
\vspace{-0.2cm}
\end{table}

\begin{table}[!htp]
\centering
\caption{effects of different trajectory proposal settings}
\vspace{-0.2cm}
\hspace{-0.2cm}
\scalebox{0.70}{
\renewcommand{\arraystretch}{1.20}
\setlength{\tabcolsep}{2.0pt}{
\begin{threeparttable}
\begin{tabular}{l|l|ccccc}
\toprule
\makecell[c]{Class} &
\makecell[c]{Method} &
minADE $(\downarrow)$ &
minFDE $(\downarrow)$ &
Miss Rate $(\downarrow)$ &
mAP $(\uparrow)$ &
Soft mAP $(\uparrow)$ 
\\ 
\hline
\multirow{3}{*}{Vehicle}
& JAM ($Y_m$=1,$K_m$=64)&  0.9036& \textbf{1.9486}& \textbf{0.3459}&\textbf{0.2153}&\textbf{0.2566}\\
& JAM ($Y_m$=8,$K_m$=8)&  0.9148& 1.9794& 0.3521&0.2023&0.2471\\
& JAM ($Y_m$=64,$K_m$=1)&  \textbf{0.9033}& 1.9585& 0.3492&0.2119&0.2551\\
\hline
\multirow{3}{*}{Pedestrian}
& JAM ($Y_m$=1,$K_m$=64)&  0.7037& 1.5041& \textbf{0.4241}&0.1372&0.1534\\
& JAM ($Y_m$=8,$K_m$=8)&  0.7071& 1.5046& 0.4299&0.1477&0.1641\\
& JAM ($Y_m$=64,$K_m$=1)&  \textbf{0.6972}& \textbf{1.4961}& 0.4278&\textbf{0.1600}&\textbf{0.1721}\\
\hline
\multirow{3}{*}{Cyclist}
& JAM ($Y_m$=1,$K_m$=64)&  1.0132& 2.2641& 0.5566&0.0839&0.0903\\
& JAM ($Y_m$=8,$K_m$=8)&  1.0201& 2.2789& 0.5528&0.0826&0.0887\\
& JAM ($Y_m$=64,$K_m$=1)&  \textbf{1.0080}& \textbf{2.2538}& \textbf{0.5524}&0\textbf{.0882}&\textbf{0.0954}\\
\hline
\multirow{3}{*}{All(Avg)}
& JAM ($Y_m$=1,$K_m$=64)&  0.8735& 1.9056& \textbf{0.4422}&0.1455&0.1668\\
& JAM ($Y_m$=8,$K_m$=8)&  0.8807& 1.9209 & 0.4449&0.1442&0.1666\\
& JAM ($Y_m$=64,$K_m$=1)& \textbf{0.8695}& \textbf{1.9028}& 0.4431& \textbf{0.1534}&\textbf{0.1742}\\
\bottomrule
\end{tabular}
\begin{tablenotes}
\footnotesize
\item We only adjusted the mode query in the proposal phase of JAM. $Y_m$ denotes the number of categories for the mode query, and $K_m$ represents the number of mode queries per category.
\end{tablenotes}
\end{threeparttable}
\label{table:ablation_proposal}
    } 
}
\vspace{-0.2cm}
\end{table}

\subsection{Comparison of Result for Different Framework}
To further demonstrate the effectiveness of the JAM framework, we conduct comparative experiments on different model frameworks. As shown in Table \ref{table:framework_cmp}, our method achieves optimal performance across nearly all metrics. For the Marginal framework setup, we find that using the classification-aware mode query design facilitates better model convergence. In the Marginal (classification-free) setup, we use 64 learnable mode queries, while in the Marginal (classification-aware) setup, we set $Y_m=8$ and $K_m=3$. For the Joint (one-step re-prediction) setup, we replace the marginal prediction in JAM with joint prediction and use six learnable mode queries. Additionally, we replicate the Joint (GameFormer) version based on the official open-source implementation of GameFormer \cite{huang2023gameformer}, which is a classic joint prediction framework that uses multi-step iteration. The results show that joint prediction models tend to overlook low probability modes, leading to poorer multi-modal performance, while marginal prediction models exhibit better multi-modal performance. However, the marginal prediction model shows the highest position error and miss rate due to its limited interaction modeling capability, while the joint prediction model excels in these metrics due to superior interaction modeling. Our proposed JAM model ($Y_m=64, K_m=1$) outperforms all other models across all metrics, indicating its effective integration of the advantages of both frameworks. Notably, we achieve better performance than GameFormer with fewer parameters. 

Additionally, we perform an ablation study on the parameter settings of the classification-aware mode query, as shown in Table \ref{table:ablation_proposal}. When $Y_m=1$, the model fully utilizes learnable mode queries. When the number of categories $Y_m=8$, we categorize based on behavior types. In contrast, when $Y_m=64$, we categorize based on clustering intention points. We find that the experimental setting with $Y_m=8$ results in a slight performance decrease compared to the unclassified method, but when the classification is sufficiently fine-grained (e.g., $Y_m=64$), the performance improves. Moreover, we observe that the use of classification-aware mode queries provides more significant gains in more irregular categories, such as pedestrians and cyclists.

\subsection{Ablation Study}
As shown in Table \ref{table:ablation_component}, we further conduct an ablation study to evaluate the contributions of the keypoint-guided encoding and classification-aware proposal mechanisms. The results indicate that the keypoint-guided encoding mechanism achieves slight improvements across various metrics, which can be attributed to the fixed target points (3s, 5s, and 8s) that guide the joint prediction module in considering both short-term and long-term predictions when handling interactions. Additionally, the improvement from the classification-aware marginal proposal mechanism is more significant, as it enables the joint prediction module in JAM to receive more comprehensive prediction modes, thereby enhancing the model's performance.

\begin{table}[!htp] 
\centering
\caption{effects of different components in JAM framework}
\vspace{-0.2cm}
\hspace{-0.2cm}
\scalebox{0.75}{
\renewcommand{\arraystretch}{1.20}
\setlength{\tabcolsep}{4.0pt}{
\begin{threeparttable}
\begin{tabular}{cc|cccc}
\toprule
\makecell[c]{keypoint-guided\\encoding} &
\makecell[c]{classification-aware\\marginal proposal} &
minADE $(\downarrow)$ &
minFDE $(\downarrow)$ &
Miss Rate $(\downarrow)$ &
mAP $(\uparrow)$ 
\\
\hline
$\times$& $\times$& 0.9192& 2.0273& 0.4764&0.1214\\
$\sqrt{}$& $\times$& 0.9047& 1.9817& 0.4675&0.1291\\
$\sqrt{}$& $\sqrt{}$& \textbf{0.8695}& \textbf{1.9028}& \textbf{0.4431}& \textbf{0.1534}\\
\bottomrule
\end{tabular}
\end{threeparttable}
\label{table:ablation_component}
    } 
}
\end{table}

\subsection{Qualitative Results}
\begin{figure}[t]
    \vspace{0.1cm}
    \centering
    \setlength{\abovecaptionskip}{0cm}
    \includegraphics[height=0.95\columnwidth]{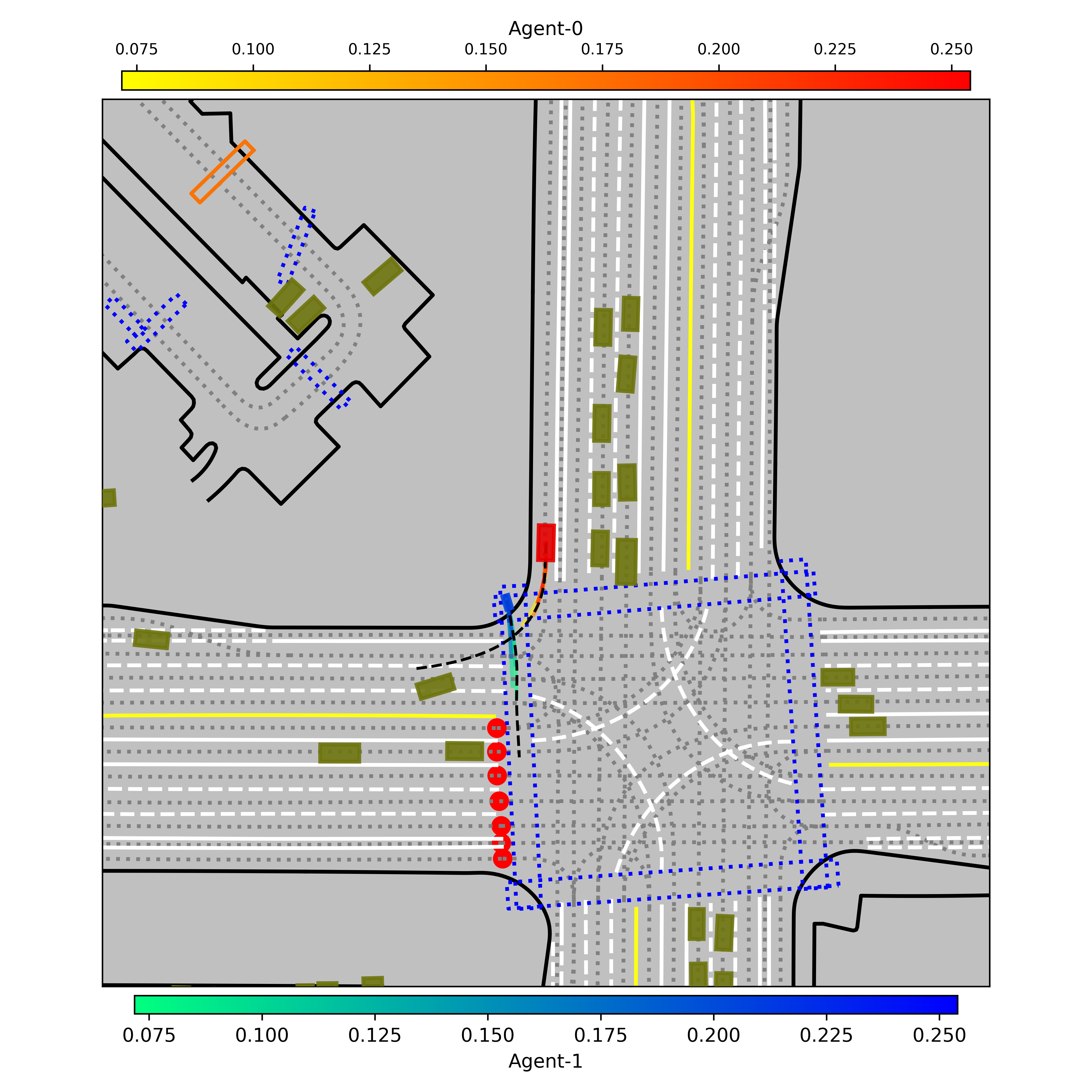}   
    \vspace{-0.1cm}
    \\
    \includegraphics[height=0.95\columnwidth]{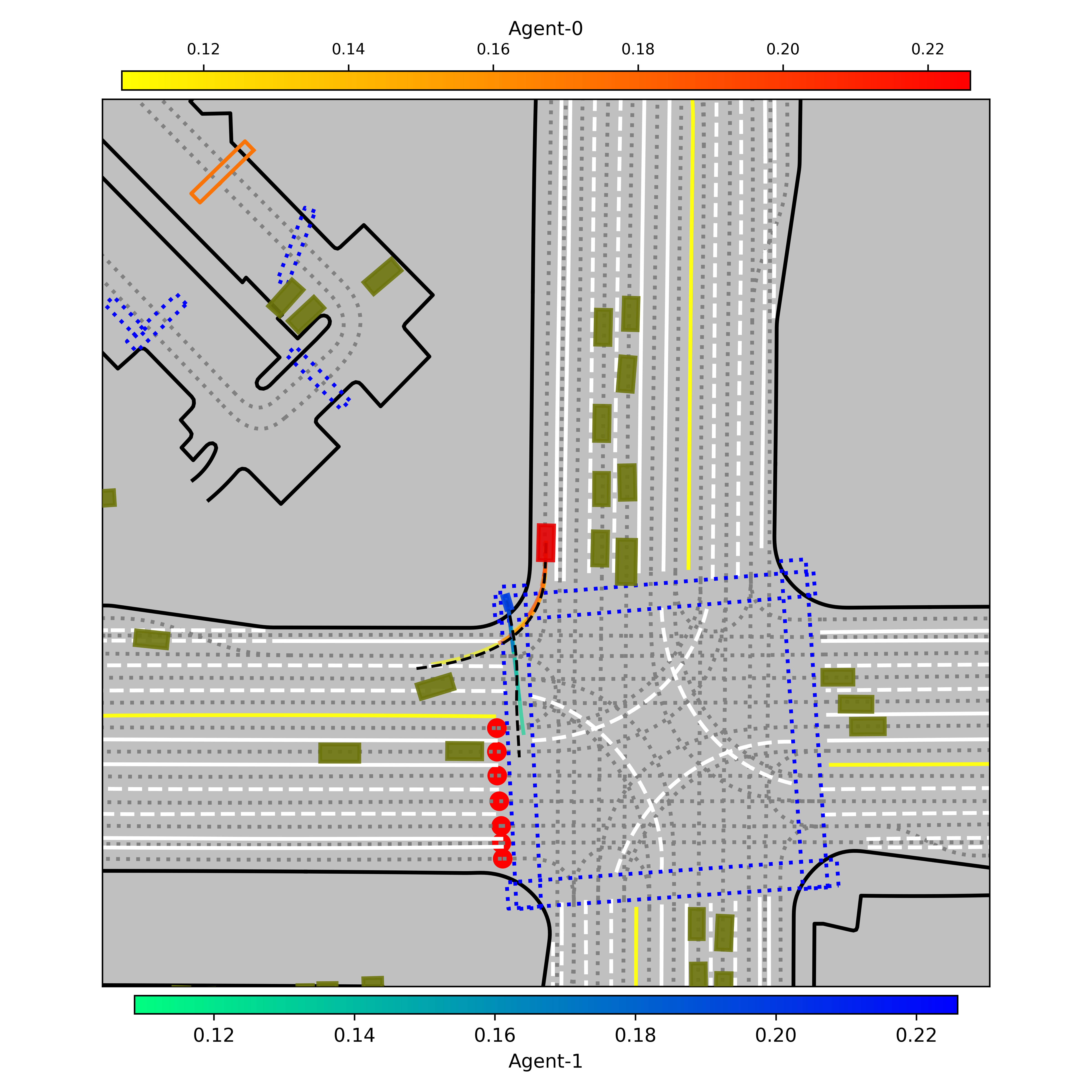}     
    \vspace{-0.3cm}
    \caption{The two-stage prediction visualization results are shown. The top image represents the marginal prediction outcomes, while the bottom image displays the trajectory results after joint prediction.}
    \vspace{-0.3cm}
    \label{fig:visual_two_stage}
\end{figure}

We evaluate the three models listed in Table \ref{table:framework_cmp} and present qualitative results in Figure \ref{fig:visual_framework_cmp}. The sharp left turns shown in the figure represent low-probability modes in the dataset. The joint prediction model with classification-free mode queries fails to accurately predict such trajectories. In contrast, both the marginal prediction model and the JAM model with classification-aware mode queries handle these cases more effectively. This is because classification-aware queries encourage the model to learn all types of motion modes, leading to better performance when encountering rare cases such as those illustrated in the figure. Furthermore, as shown in Figure \ref{fig:visual_two_stage}, the refined trajectories from the joint prediction module are closer to the ground truth. This observation demonstrates the effectiveness of our proposed approach.

\section{Conclusion}
In this paper, we propose a two-stage interactive prediction framework that combines marginal prediction with joint prediction. This framework aims to address the challenge of low-quality generation of low-probability modes in joint prediction. We classify trajectories and explicitly encourage the model to learn and generate all possible modes to address this problem. Additionally, we introduce a keypoint-guided joint prediction module to better extract information from the proposals. Experimental results demonstrate that the JAM framework achieves competitive performance in interactive trajectory prediction.

In future work, we believe that extending the proposed classification-aware prediction method to joint prediction is a promising direction.  In addition, we consider it interesting to further explore the keypoint-guided mechanism, for example, by using interactive keypoint pairs instead of only stage-wise target points.

\bibliographystyle{IEEEtran}
\bibliography{main}

\end{document}